\documentclass[11pt]{article}

\usepackage[final]{acl}
\usepackage{times}
\usepackage{latexsym}
\usepackage[T1]{fontenc}
\usepackage[utf8]{inputenc}
\usepackage{microtype}
\usepackage{inconsolata}
\usepackage{amsmath}
\usepackage{amssymb}
\usepackage{booktabs}
\usepackage{array}
\usepackage{multirow}
\usepackage{graphicx}
\usepackage{enumitem}
\usepackage{xcolor}
\usepackage{url}

\newcommand{\method}{$\Delta$-MOPD}
\newcommand{\sg}{\operatorname{sg}}

\title{Composing What Each Teacher Learned: Delta Distillation for Multi-Teacher On-Policy Learning}

\author{
  \textbf{Yuanda Xu}\thanks{Equal contribution.}\,
  \thanks{Correspondence to \texttt{yuanda@math.princeton.edu}} \quad
  \textbf{Hejian Sang}\textsuperscript{2}\footnotemark[1] \quad
  \textbf{Zhengze Zhou}\textsuperscript{1}\footnotemark[1] \quad
  \textbf{Shayan Mohajer Hamidi} \quad
  \\[0.5em]
  \textbf{Xiaomin Li}\textsuperscript{3} \quad
  \textbf{Rohit Jain}\textsuperscript{1} \quad
  \textbf{Alborz Geramifard}\textsuperscript{1}
  \\[0.5em]
  \textsuperscript{1}LinkedIn \quad
  \textsuperscript{2}Iowa State University \quad
  \textsuperscript{3}Harvard University
}
\begin{document}
\maketitle
\begin{abstract}
Multi-teacher on-policy distillation (MOPD) is used in two settings. In
\emph{common-domain composition}, several teachers score each student
rollout from one prompt domain and their signals form a single target; in
\emph{routed-domain distillation}, prompts from different domains are
assigned to the corresponding specialist. Both settings usually transfer
each teacher's endpoint policy, which mixes what post-training changed
with preferences inherited from the teacher's base. We introduce
\method{}, which transfers each teacher's teacher-minus-base logit shift
re-anchored at the student's frozen initialization, and compare it with
endpoint supervision in both settings while holding teacher selection
fixed. We first expose the mechanism that impedes endpoint transfer:
inherited base pull can exceed the post-training shift. Removing it
reduces the teacher-term norm ratio from $5.2\!:\!1$ to $1.44\!:\!1$ and
target--student KL from $0.152$ to $0.031$. The resulting target reaches
the endpoint arm's best evaluated Math score with $35\%$ fewer allocated
H100-hours.
Across our experiments, the results suggest that shift targets are
particularly useful when teacher signals are combined at a state. With
three composed teachers, \method{} exceeds endpoint composition by
$4.11$ Math and $1.95$ five-benchmark points; with two, it matches
endpoint accuracy. In an unpaired cross-tokenizer extension, a projected
fourth shift improves all three Math benchmarks and adds a further
$3.32$ Math and $1.85$ five-benchmark points. Under phased routing, it achieves higher mean
performance in both phase orders and reduces the observed order gap from
$10.50$ to $6.42$ points. Under interleaved routing, where each update
involves one teacher, the two targets perform comparably. The phased
results provide supporting evidence that the benefit may extend to
signals accumulated across training phases. Target construction is thus
an independent design axis in MOPD, complementary to teacher selection.
The experiment repo could be find at  \url{https://github.com/HJSang/Delta-MOPD}
\end{abstract}

\section{Introduction}

On-policy distillation (OPD) trains a student on its own rollouts using
scores from a frozen teacher, so supervision arrives at the states the
student visits \citep{gu2024minillm,agarwal2024onpolicy,song2026opdsurvey}.
Because post-trained specialists are individually incomplete and widely
released, multi-teacher OPD (MOPD) aims to combine several of them in one
student. MOPD is used in two settings. In \emph{common-domain
composition}, several teachers score the same rollout from one prompt
domain and their signals are merged into one target. In
\emph{routed-domain distillation}, prompts from different domains are
assigned to the corresponding specialist, so each state is supervised by
one teacher.

Both settings, and the studies of routing, teacher counteraction, and
cascaded overwrite built on them
\citep{ma2026mopdmultiteacheronpolicydistillation,
chen2026counteractionawaremultiteacheronpolicydistillation,
li2026everycoin}, transfer each teacher's \emph{endpoint} policy. A
specialist endpoint, however, combines the change produced by its
post-training stage with preferences inherited from its base
\citep{deepseek2025r1}. When a teacher's base differs from the student's
initialization, endpoint supervision transfers that inherited difference
as well. The original MOPD pipeline avoids it because every teacher shares
the student's initial checkpoint
\citep{ma2026mopdmultiteacheronpolicydistillation}; public specialists
rarely do. Single-teacher methods report promising results from
teacher-relative differences
\citep{heo2026onpolicydeltadistillation,
feng2026weaktostronggeneralizationdirectonpolicy,
yang2026learningteachergeneralizedonpolicy}, but whether such differences
are a better transfer object in either MOPD setting has not been tested.

\begin{figure*}[t]
\centering
\includegraphics[width=\textwidth]{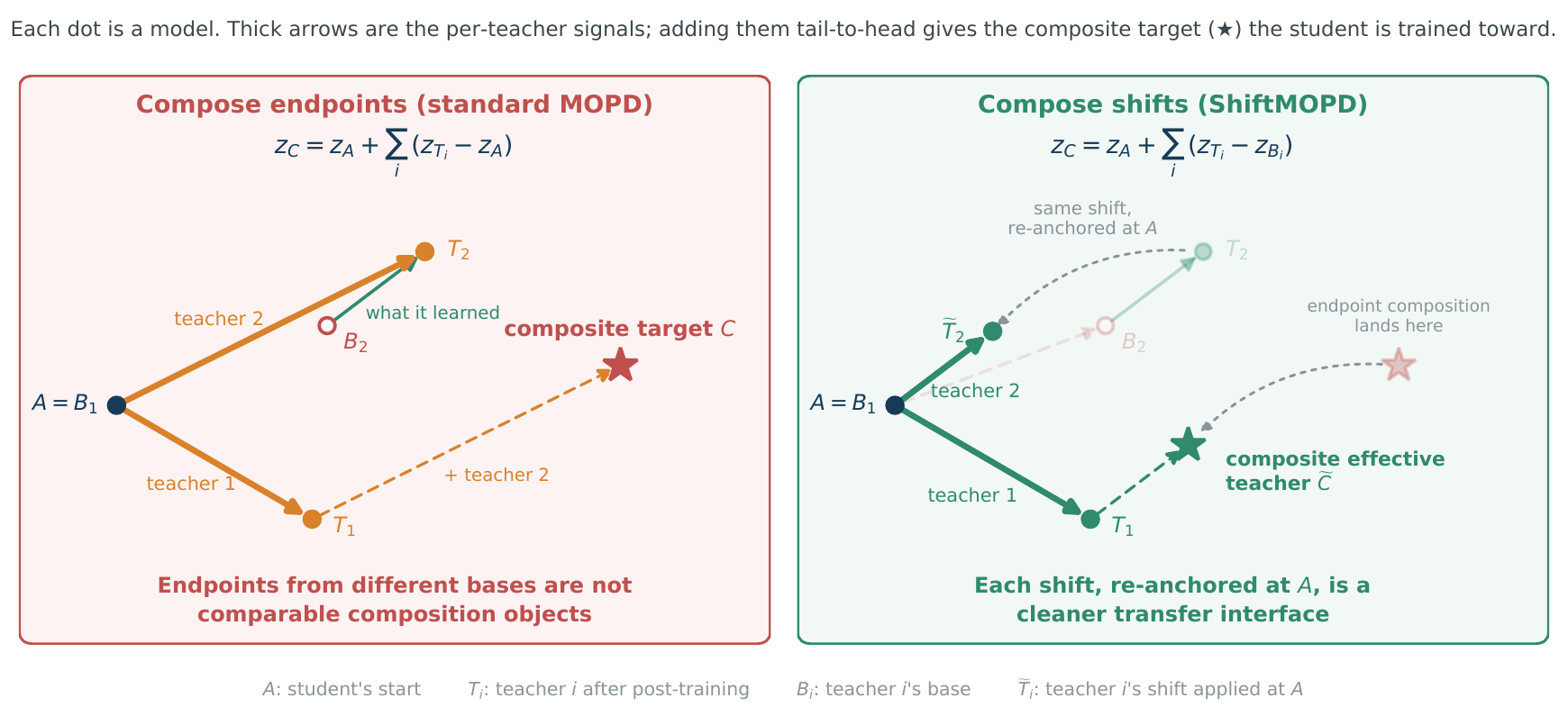}
\caption{Endpoint versus shift targets. Each endpoint term $z_{T_i}-z_A$
contains a post-training shift $z_{T_i}-z_{B_i}$ and an inherited base
difference $z_{B_i}-z_A$. \method\ keeps only the shift and re-anchors it
at $A$, giving the effective teacher $\widetilde{T}_i$; the two composite
targets differ by the summed base differences. Teacher~1 is drawn with
$B_1=A$, so $\widetilde{T}_1=T_1$. Positions are schematic.}
\label{fig:overview}
\end{figure*}

We separate two decisions in any MOPD recipe: \emph{teacher selection},
which teachers supervise a state, and \emph{target construction}, how
their scores define the distribution to be matched. Holding selection
fixed, we compare endpoint targets with \method{}, which measures each
teacher against its own base $B_i$ and re-anchors the resulting shift at
the student's initialization $A$. Writing $z_X$ for checkpoint $X$'s
logits at a student-generated state and $\mathcal S$ for the selected
teachers, the two targets are
\begin{equation}
\begin{aligned}
\text{endpoint:}\quad z_{\mathrm E}
&=z_A+\sum\nolimits_{i\in\mathcal S}(z_{T_i}-z_A),\\
\text{\method:}\quad z_C
&=z_A+\sum\nolimits_{i\in\mathcal S}(z_{T_i}-z_{B_i}),
\end{aligned}
\label{eq:intro-contrast}
\end{equation}
each normalized by a softmax. Composition selects all teachers; routing
selects one. When $B_i=A$, both constructions give the same term, so only
cross-origin teachers are affected (Figure~\ref{fig:overview}).
Concretely, if a Math specialist both acquires stronger reasoning and
keeps a verbosity preference already present in its base, endpoint
distillation transfers both; dividing the specialist by its base cancels
what the base already preferred, and the anchor expresses the remaining
change relative to the student's own starting policy. We hypothesize that
removing inherited base differences yields a target that is closer to the
student and, under composition, more balanced across teachers.

Our results suggest that shift targets are particularly useful when
teacher signals are combined at a state. The mechanism appears directly
in the learning signal: endpoint supervision gives the inherited base
pull a larger gradient than the post-training shift, whereas base
subtraction produces a target that is five times closer to the student
and substantially more balanced across teachers. The closer target
reaches the endpoint arm's best Math score with $35\%$ fewer allocated
H100-hours. In common-domain composition, shifts match endpoint accuracy
with two teachers and lead on Math and the five-benchmark macro with
three. In an unpaired cross-tokenizer extension, a projected fourth shift
from a second cross-origin teacher improves all three Math benchmarks,
raising the Math macro by a further $3.32$ points and the five-benchmark
macro by $1.85$ points. In routed-domain distillation, phased routing yields higher mean
performance for shifts in both phase orders and a smaller observed order
gap; interleaved routing, where each update sees a single teacher, yields
comparable results. The phased results provide supporting evidence that
the benefit may extend to signals accumulated across training phases.

\paragraph{Contributions.}
\begin{enumerate}[leftmargin=1.5em,itemsep=2pt,topsep=2pt]
  \item We identify target construction as a design axis in MOPD,
  independent of teacher selection, and instantiate it with \method{}
  (Eq.~\ref{eq:intro-contrast}), which reduces exactly to endpoint OPD
  for same-origin teachers.
  \item \textbf{We expose the mechanism that impedes endpoint transfer.}
  The inherited base pull can exceed the post-training shift; removing it
  lowers the teacher-term norm ratio from $5.2\!:\!1$ to $1.44\!:\!1$,
  and reduces target--student KL fivefold.
  \item \textbf{The improved geometry yields stronger signal
  combination.} With three composed teachers, \method{} exceeds endpoint
  composition by $4.11$ Math and $1.95$ five-benchmark points; under
  an unpaired cross-tokenizer extension, a projected fourth shift improves
  all three Math benchmarks and adds a further $3.32$ Math and $1.85$
  five-benchmark points. Under phased
  routing it achieves higher mean performance in both orders and reduces
  the observed order gap from $10.50$ to $6.42$ points. Interleaved
  routing performs comparably.
\end{enumerate}

\section{Preliminaries}
\label{sec:prelim}

Let $x\sim\mathcal{D}$ and
$y=(y_1,\ldots,y_L)\sim\pi_\theta(\cdot\mid x)$, with
$h_t=(x,y_{<t})$, $p_t=\pi_\theta(\cdot\mid h_t)$, and $q_t$ the target
distribution at $h_t$. OPD minimizes the token-level reverse KL
\begin{equation}
\begin{aligned}
\mathcal{L}
={}&\mathbb{E}_{x\sim\mathcal D,\;y\sim\pi_\theta(\cdot\mid x)}\\
&\left[
\frac{1}{L}\sum_{t=1}^{L}
D_{\mathrm{KL}}\!\left(p_t\,\|\,q_t\right)
\right].
\end{aligned}
\label{eq:endpoint-loss}
\end{equation}
For sampled $y_t\sim p_t$, $\log p_t(y_t)-\log q_t(y_t)$ is a one-sample
estimator of the per-state reverse KL. Training maximizes the
sampled-token reward $\log q_t(y_t)-\log p_t(y_t)$ with the centered
score-function estimator of Appendix~\ref{app:diagnostic-definitions}.

\paragraph{Notation.}
$\pi_X$ and $z_X$ denote checkpoint $X$'s distribution and logits over a
shared effective vocabulary. $A$ is a frozen anchor, in every experiment
the student initialization; $T_i$ is teacher $i$'s post-training
endpoint; $B_i$ is the exact checkpoint that entered that stage. The
\emph{shift} of teacher $i$ is $\delta_i=z_{T_i}-z_{B_i}$ and its
\emph{inherited base difference} is $z_{B_i}-z_A$. A shift carries the
complete effect of post-training---capability, style, and calibration
alike---and is not assumed to isolate capability. Teacher $i$ is
\emph{same-origin} when $B_i=A$ and \emph{cross-origin} otherwise.

\paragraph{Teacher selection.}
A selection rule assigns each state a teacher set $\mathcal S(h_t)$.
Common-domain composition uses $\mathcal S(h_t)=\{1,\ldots,M\}$, so the
teachers' relative magnitudes at a state determine the update.
Routed-domain distillation uses $\mathcal S(h_t)=\{\rho(h_t)\}$, where
$\rho$ maps a prompt to its domain's teacher; domains may be interleaved
within training or assigned to contiguous phases.

\section{\method}
\label{sec:method}

\subsection{Construction}

For teachers that share a token-id mapping on the $151{,}665$-token
effective vocabulary, \method\ uses the shared-anchor target
\begin{equation}
\pi_C(v\mid h_t)
=\frac{1}{Z(h_t)}
\pi_A(v\mid h_t)
\prod_{i\in\mathcal S(h_t)}
\frac{\pi_{T_i}(v\mid h_t)}{\pi_{B_i}(v\mid h_t)},
\label{eq:composite-ratio}
\end{equation}
where $Z(h_t)$ normalizes over the vocabulary. Each model's
log-partition term is constant across tokens, so the target is a logit
sum normalized once:
\begin{equation}
\begin{aligned}
z_C(v\mid h_t)
&=z_A(v\mid h_t)+\sum_{i\in\mathcal S(h_t)}\delta_i(v\mid h_t),\\
\pi_C(\cdot\mid h_t)
&=\operatorname{softmax}\!\big(z_C(\cdot\mid h_t)\big),
\end{aligned}
\label{eq:composite-logits}
\end{equation}
with sampled-token reward
\begin{equation}
R^C_t(v)
=\log\pi_C(v\mid h_t)-\log\pi_\theta(v\mid h_t).
\label{eq:composite-reward}
\end{equation}
Setting $q_t=\pi_C(\cdot\mid h_t)$ in Eq.~\ref{eq:endpoint-loss} gives the
full-vocabulary objective. Some runs approximate only the partition
function on a restricted support; details of this approximation are
given in Appendix~\ref{app:diagnostic-definitions}.

\paragraph{Endpoint control.}
The control replaces each shift by $z_{T_i}-z_A$, giving
$z_{\mathrm E}$ in Eq.~\ref{eq:intro-contrast}. At a given prefix the two
pre-softmax targets differ by exactly
$\sum_{i\in\mathcal S(h_t)}(z_{B_i}-z_A)$. With one selected teacher the
control is that teacher's endpoint.

\paragraph{Same-origin reduction.}
If $B_i=A$, then $z_A+\delta_i=z_{T_i}$: a same-origin teacher's shift
target \emph{is} its endpoint. Routing to a same-origin teacher is
therefore exactly endpoint OPD, and if every selected teacher is
same-origin the two composites coincide. Only cross-origin terms differ.

\subsection{Reference-term interpretation}
\label{sec:decomposition}

The endpoint reward of teacher $j$ decomposes exactly as
\begin{equation}
\begin{aligned}
\log\pi_{T_j} - \log\pi_\theta
={}&
\underbrace{\log\pi_{T_j} - \log\pi_{B_j}}_{R^\Delta_j:\ \text{shift}}\\
&+
\underbrace{\log\pi_{B_j} - \log\pi_\theta}_{R^B_j:\
\text{base-reference pull}}.
\end{aligned}
\label{eq:decomposition}
\end{equation}
The base-reference pull equals the inherited difference
$\log\pi_{B_j}-\log\pi_A$ plus the current anchor--student gap
$\log\pi_A-\log\pi_\theta$; at initialization only the inherited
difference remains. Up to a token-independent normalizer, \method\ has
the same form with a single shared reference:
\begin{equation*}
\begin{aligned}
R^C_t(v)={}&\sum_{i\in\mathcal S(h_t)}\delta_i(v\mid h_t)\\
&+\big[\log\pi_A(v\mid h_t)-\log\pi_\theta(v\mid h_t)\big].
\end{aligned}
\end{equation*}
Both targets reward following teacher shifts and pull toward a reference.
Endpoint supervision pulls toward each selected teacher's own base;
\method\ replaces these heterogeneous pulls by one pull toward the
student's initialization, which is zero at the start of training. Two
consequences are plausible: the target is closer to the student, and,
under composition, a cross-origin teacher's inherited difference no
longer sets the scale of its term relative to the others.
Section~\ref{sec:diagnosis} examines both.

\section{Experimental Setup}
\label{sec:experimental-setup}

\paragraph{Checkpoints.}
The anchor and student initialization in every run is
DeepSeek-R1-Distill-Qwen-1.5B \citep{deepseek2025r1}. Table~\ref{tab:setup}
lists the teachers and the exact precursor used for each shift.
Polaris-7B is post-trained from the 7B checkpoint of the same
distillation family, so ``cross-origin'' here means a different and
larger base of the same lineage. Nemotron and JustRL are same-origin, so
their shift and endpoint terms are identical. In every paired comparison,
the two targets therefore differ by one controlled substitution: the
endpoint or teacher-relative representation of Polaris. A
tokenizer-projected Qwen3-DAPO-449 shift extends the construction to a
second cross-origin teacher in
Appendix~\ref{app:qwen3}.

\begin{table*}[t]
\centering
\small
\caption{Public checkpoints. ``Base'' is the exact precursor used to
compute each shift. Nemotron is pinned to revision \texttt{v1}.}
\label{tab:setup}
\resizebox{\textwidth}{!}{%
\begin{tabular}{llllll}
\toprule
Role & Model & Base & Origin & Specialization & Use \\
\midrule
Anchor / student init & DS-R1-Distill-Qwen-1.5B & self & self & -- & all runs\\
Math teacher & JustRL-DeepSeek-1.5B & DS-R1-Distill-Qwen-1.5B
& same & Math RL & composition ($M=3$)\\
Science/IF teacher & Nemotron-Reasoning-Qwen-1.5B & DS-R1-Distill-Qwen-1.5B
& same & multi-domain RL & composition, routing\\
Math teacher & Polaris-7B-Preview & DS-R1-Distill-Qwen-7B
& cross & Math RL & all runs\\
Math teacher & Qwen3-DAPO-449 & Qwen3-8B
& cross & DAPO Math RL & extension (App.~\ref{app:qwen3})\\
\bottomrule
\end{tabular}
}
\end{table*}

\paragraph{Runs.}
\emph{Common-domain composition} uses two runs on BigMath prompts, with
every teacher scoring every rollout: the \textbf{mechanism} run composes
Nemotron and Polaris ($M=2$) and records diagnostics; the
\textbf{scaling} run compares $M=2$ with $M=3$, which adds JustRL.
\emph{Routed-domain distillation} assigns Math prompts to Polaris and
Science/IF prompts to Nemotron, either \textbf{interleaved} within each
batch or in two \textbf{phased} orders. The \textbf{acquisition} run
distills Polaris alone. Paired arms share initialization, prompts,
optimizer, update budget, and evaluation; all frozen models score the
same prefixes within an arm, although the arms' on-policy trajectories
diverge. Each comparison uses a shared run protocol; key configurations
are listed in Table~\ref{tab:as-run-training}.

\subsection{Diagnostic metrics}
\label{sec:diagnostics}

We measure $\Gamma_{\mathrm{base}}=\|g_B\|/\|g_\Delta\|$, the gradient
norm of the base-reference pull relative to the shift. For composition
geometry, we report the teacher-term norm ratio, centered-shift cosine,
cancellation, and top-$16$ sign conflict. Target--student and
target--anchor KL are computed over the full effective vocabulary,
independently of the training normalizer. The reported forward KL
$D_{\mathrm{KL}}(\pi_C\|\pi_\theta)$ measures target tracking and is not
the training objective. Definitions are in
Appendix~\ref{app:diagnostic-definitions}.

\section{Results}
\label{sec:results}

The evidence follows a mechanism-to-outcome progression. We first test
whether inherited base differences distort the learning signal
(\S\ref{sec:diagnosis}), then examine the predicted consequences for
cross-origin acquisition (\S\ref{sec:cross-origin}), common-domain
composition (\S\ref{sec:composition}), and routed-domain distillation
(\S\ref{sec:routing}).

\subsection{Inherited base differences distort composition geometry}
\label{sec:diagnosis}

\begin{figure*}[t]
\centering
\includegraphics[width=\textwidth]{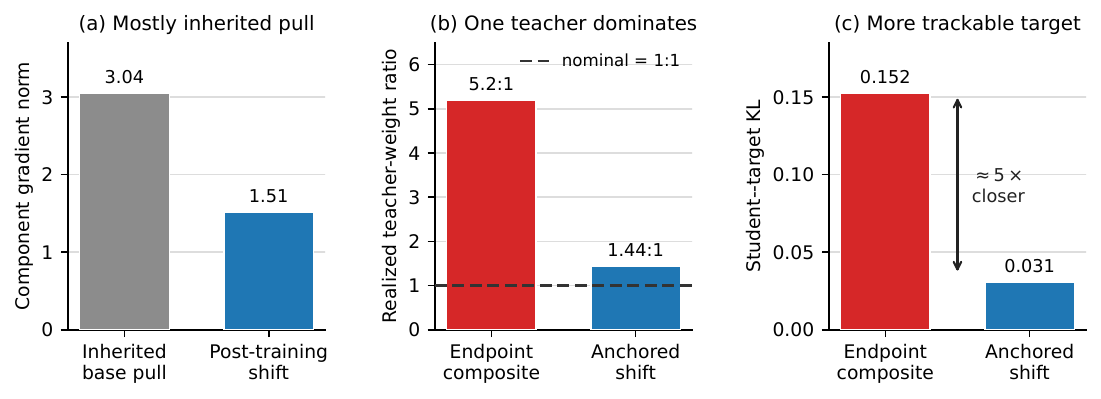}
\caption{Composition diagnostics in the mechanism run.
\textbf{(a)} For Polaris, the base-reference pull has a larger gradient
norm than the shift. \textbf{(b)} The endpoint composite has a
$5.2\!:\!1$ teacher-term norm ratio; the shift composite has $1.44\!:\!1$.
\textbf{(c)} The shift target is about five times closer to the student
in target--student KL. Values are late-training averages
(Table~\ref{tab:k2-composition-diagnostics}).}
\label{fig:hero}
\end{figure*}

\paragraph{Inherited pull dominates early learning.}
Under endpoint supervision, the base-reference pull of Polaris has
roughly twice the gradient norm of its shift
($\Gamma_{\mathrm{base}}\!\approx\!2$, averaged over checkpoints recorded
every $20$ steps). The ratio declines during training because only
$R^B_j$ depends on the student: it shrinks as the student moves toward
$\pi_{B_j}$, whereas $R^\Delta_j$ is fixed by the teacher--base pair at a
given state. The inherited component therefore has its greatest relative
influence while the specialization is still being acquired.

\paragraph{Base subtraction restores balance.}
With the Nemotron term identical in both arms, the teacher-term norm
ratio falls from $5.2$ under endpoint composition to $1.44$ under shift
composition, and target--student KL falls from $0.152$ to $0.031$
(Figure~\ref{fig:hero}). The shift composite lies near the equal-norm
independence reference (cancellation $0.287$ versus $0.293$, cosine
$-0.016$), whereas the endpoint composite has low cancellation ($0.136$),
the signature of one dominant term. The ordering appears by step $20$ and
persists through training (Appendix~\ref{app:composition-details}); at
$M=3$, target--anchor KL remains lower for shifts ($0.483$ versus
$0.695$).

Together, these measurements establish the proposed mechanism: inherited
base differences enlarge the target and set the relative scale of the
teacher terms. Removing them yields two testable predictions. The closer
target should be acquired earlier, and the more balanced terms should
make additional teachers more useful. We test these predictions next.

\subsection{A closer target accelerates cross-origin acquisition}
\label{sec:cross-origin}
\label{sec:learnability}

Distilling Polaris alone removes composition and directly tests the first
prediction. \method{} leads the English-Math macro by $2.87$ pp at step
$100$, after which the arms approach a similar level. Including
precursor-scoring overhead, \method{} scores $47.90\%$ at step $100$,
already exceeding Endpoint's best evaluated $46.54\%$ at step $195$;
after charging for \method{}'s slower updates, these checkpoints cost
$28.0$ versus $42.9$ allocated H100-hours. At initialization, the
endpoint target is
$4.94\times$ farther from the student in target--student KL across
$16{,}384$ shared prefix states. The closer target is therefore acquired
earlier even after charging for the additional precursor forward pass.
Details are in Appendix~\ref{app:k1-later}.

\subsection{Balanced shifts improve common-domain composition}
\label{sec:composition}

We next test the second prediction: whether balancing the teacher terms
turns additional supervision into usable capability.

\paragraph{Two teachers.}
In the mechanism run, both composites improve substantially over the
initial student and reach similar aggregate accuracy: $44.91\%$ for
\method\ versus $45.14\%$ for the endpoint composite under Avg@$K$, and
$37.73\%$ versus $36.25\%$ under greedy decoding. \method\ is higher on
Science/IF under both decodings, while the endpoint composite is
$3.4$ pp higher on MATH-500 (Appendix~\ref{app:composition-details}).

\paragraph{Three teachers.}
The scaling run compares $M=2$ with $M=3$ under one protocol
(Table~\ref{tab:composition}). At $M=3$, \method\ exceeds the endpoint
composite by $4.11$ pp on the Math macro and $1.95$ pp on the five-suite
macro. Adding the same JustRL teacher raises shift composition by
$4.40$ Math and $1.78$ five-suite pp, but endpoint composition by only
$0.43$ and $0.58$. Because JustRL is same-origin, its term is identical in
both arms; the difference lies in how Polaris is represented when a third
teacher is added. The gain is concentrated in AIME 2025; Science/IF,
which receives no in-domain prompts in this run, is $1.29$ pp below the
endpoint composite (Table~\ref{tab:m3-full-capability}).

\begin{table}[t]
\centering
\small
\caption{Common-domain composition in the scaling run at step $100$.
Greedy pass@$1$ macros (\%). All teachers score every BigMath rollout.
$\Delta_{3-2}$ is the change from adding same-origin JustRL. Per-benchmark
results are in Table~\ref{tab:m3-full-capability}.}
\label{tab:composition}
\setlength{\tabcolsep}{3.5pt}
\resizebox{\columnwidth}{!}{%
\begin{tabular}{@{}llccc@{}}
\toprule
Teachers & Target & Math & Sci/IF & Five-suite\\
\midrule
$M=2$ & Endpoint composite & 28.96 & 15.70 & 23.66\\
$M=2$ & \method & 29.10 & \textbf{17.38} & 24.41\\
$M=3$ & Endpoint composite & 29.39 & 16.51 & 24.24\\
$M=3$ & \method & \textbf{33.50} & 15.22 & \textbf{26.19}\\
\midrule
$\Delta_{3-2}$ & Endpoint composite & $+0.43$ & $+0.81$ & $+0.58$\\
$\Delta_{3-2}$ & \method & $\mathbf{+4.40}$ & $-2.16$ & $\mathbf{+1.78}$\\
\bottomrule
\end{tabular}
}
\end{table}

Shift composition thus preserves two-teacher accuracy and turns an
additional teacher into a substantially larger Math gain.

\subsection{Shift targets show consistent gains under phased routing}
\label{sec:routing}

Composition combines teacher signals within each state. Routing separates
them across examples or training phases, examining whether the same target
construction also changes how specialist signals accumulate over time.
Under routing, Polaris supervises Math prompts and Nemotron supervises
Science/IF prompts. Because Nemotron is same-origin, the two targets
coincide on Science/IF states and differ only in the routed Polaris term.
Table~\ref{tab:routing} reports two schedules.

\begin{table}[t]
\centering
\small
\caption{Routed-domain distillation. Interleaved routing mixes domains in
each batch ($50/25/25$ Math/Science/IF); phased routing trains one
teacher--domain phase at a time. Greedy pass@$1$ macros (\%); $\Delta$
is \method\ minus Endpoint in pp.}
\label{tab:routing}
\setlength{\tabcolsep}{3.5pt}
\resizebox{\columnwidth}{!}{%
\begin{tabular}{@{}llccc@{}}
\toprule
Schedule & Macro & Endpoint & \method & $\Delta$\\
\midrule
\multirow{3}{*}{Interleaved}
& Math & 30.61 & 31.52 & $+0.91$\\
& Sci/IF & 18.26 & 18.45 & $+0.19$\\
& Five-suite & 25.67 & 26.29 & $+0.62$\\
\midrule
\multirow{3}{*}{Phased: Sci/IF $\rightarrow$ Math}
& Math & 28.02 & 34.39 & $+6.37$\\
& Sci/IF & 18.25 & 22.89 & $+4.64$\\
& Five-suite & 24.11 & 29.79 & $+5.68$\\
\midrule
\multirow{3}{*}{Phased: Math $\rightarrow$ Sci/IF}
& Math & 40.87 & 42.20 & $+1.33$\\
& Sci/IF & 25.23 & 27.23 & $+2.00$\\
& Five-suite & 34.61 & 36.21 & $+1.60$\\
\bottomrule
\end{tabular}
}
\end{table}

\paragraph{Interleaved routing.}
With both teachers active throughout training, the five-suite macro
differs by only $0.62$ pp in favor of \method, with
group-macro differences of $+0.91$ pp on Math and $+0.19$ pp on
Science/IF. The two targets are therefore comparable when each update
involves a single teacher, as expected if the benefit of shifts comes
from reconciling several teacher signals.

\paragraph{Phased routing.}
\label{sec:sequential}
When each teacher--domain pair occupies a contiguous phase, \method\ ends
above its same-order endpoint control in both orders, by $5.68$ and
$1.60$ five-suite pp, and both group macros improve in each order. The
gap between the two phase orders falls from $10.50$ pp under endpoint
supervision to $6.42$ pp. Phased training couples each teacher with its
domain data, providing evidence about the robustness of the combined
teacher--data curriculum across phase orders.

Together, these results suggest that the benefits of shift targets may
extend beyond within-state composition to settings where teacher signals
accumulate across training phases. We view the phased-routing results as
supporting evidence for this interpretation, while interleaved routing
yields comparable performance.

\section{Discussion}
\label{sec:discussion}

\paragraph{Where shifts help.}
Shift targets show their clearest benefits when several teachers are
composed at a state. Phased routing provides supporting evidence that
these benefits may extend to settings where routed teachers occupy
separate phases and their signals accumulate over training; under
interleaved routing, where each update involves one teacher, the targets
are comparable. The paired design holds all same-origin terms fixed and
changes only the representation of Polaris, directly attributing the
contrast to target construction. The projected four-teacher extension
adds a second cross-origin shift, improves all three Math benchmarks, and
raises the Math macro by a further $3.32$ pp
(Appendix~\ref{app:qwen3}), extending the construction to a larger
heterogeneous teacher set.

\paragraph{Geometry and scale.}
Cosine and sign conflict are invariant to rescaling individual shifts,
and cancellation to rescaling all shifts jointly. Their change therefore
reflects the relative geometry of the teacher terms, not only the overall
distance to the student. The policy-ratio form cancels vocabulary
normalization constants but does not calibrate differences in logit
temperature or confidence across checkpoints, so shift norms reflect both
the magnitude of post-training change and the scale on which it is
expressed. Calibration-aware composition is complementary to base
subtraction.

\section{Related Work}

\paragraph{Distillation for language models.}
Knowledge distillation transfers a teacher's output distribution to a
smaller student \citep{hinton2015distilling}, for language models mostly
off-policy on teacher-written rationales
\citep{li2022explanations,hsieh2023distilling,fu2023specializing,
singh2024beyond,liu2024ddk}, so the student is never supervised at the
states it visits. OPD removes that mismatch by scoring the student's own
rollouts \citep{agarwal2024onpolicy,song2026opdsurvey} and has been
reframed as compressing reward-shaped behavior
\citep{sang2026crisp,xu2026tip,he2026selfdistillationzero,
xu2026grpoonpolicydistillationempirical}.
Work that broadens what the on-policy signal carries, through rubric
judgments \citep{fang2026rubricopd}, dual sequence- and token-level views
\citep{hou2026uniopd,wang2026tokenteachability,
shen2026inputspecificcredit}, or context rather than weights
\citep{ye2026opcd}, keeps the teacher's endpoint as the matched object.
\citet{li2026rethinkopd} show that single-teacher OPD depends on
thinking-pattern compatibility rather than teacher strength; student
competence frontiers for on-policy self-distillation are studied by
\citet{xu2026paced}.

\paragraph{Multi-teacher on-policy distillation.}
MOPD studies of routing, counteraction, and cascaded overwrite
\citep{ma2026mopdmultiteacheronpolicydistillation,
chen2026counteractionawaremultiteacheronpolicydistillation,
li2026everycoin} motivate our paired controls, and the same brittleness
surfaces in tool-use, GUI-agent, and expert-to-generalist settings
\citep{shen2026diagnosingcalibratingtoolcallboundary,
lian2026uimopdmultiplatformonpolicydistillation,
chen2026regenreplayrecyclingexperttogeneralistdistillation}.
Sample routing connects group-relative and self-distillation policy
optimization \citep{li2026samplerouting}. These methods act on
selection, scheduling, or rehearsal; we vary what each selected teacher
contributes.

\paragraph{Differences as transfer objects.}
Delta-oriented single-teacher methods use teacher--base differences for
direct or extrapolated transfer
\citep{heo2026onpolicydeltadistillation,
feng2026weaktostronggeneralizationdirectonpolicy,
yang2026learningteachergeneralizedonpolicy}, and task arithmetic and
proxy tuning compose parameter- or logit-space differences offline
\citep{ilharco2023task,liu2024proxy}. We study such differences as
transfer objects in both MOPD settings, on the student's own prefixes and
normalized under one softmax.

\section{Conclusion}

We studied target construction in MOPD by comparing teacher-relative
shifts, re-anchored at the student's initialization, with endpoint
supervision. Inherited base differences can dominate a teacher's
post-training shift; removing them yields a target that is balanced
across teachers and closer to the student. Across our experiments, shift
targets show their clearest benefits when teacher signals are combined
at a state: they match endpoints with two composed teachers and raise
Math and five-benchmark accuracy with three. In an unpaired
cross-tokenizer extension, adding a projected fourth shift from a second
cross-origin teacher yields further gains on both macros. Under
phased routing, shifts achieve higher mean performance in both tested
orders and reduce the observed order gap, providing supporting evidence
that their benefits may extend to teacher signals accumulated across
phases. Under interleaved routing, where each update sees one teacher,
the targets are comparable. Because the construction reduces exactly to
endpoint OPD for same-origin teachers, it preserves their signal and acts
only on inherited base differences. What is transferred from each
teacher is thus a design axis in MOPD, complementary to which teachers
are selected.

\section{Limitations}

\begin{itemize}[leftmargin=1.3em,itemsep=2pt,topsep=2pt]
  \item \textbf{Evaluation scope.} We study public reasoning specialists
  and common-domain composition on Math prompts, with Science/IF measured
  through held-out evaluation. Broader specialist families and
  domain-mixed composition are natural extensions.
  \item \textbf{Requirements.} \method{} requires each teacher's
  precursor and a compatible tokenizer or an explicit projection rule,
  unlike black-box OPD \citep{ye2025blackboxopd}.
\end{itemize}

\bibliography{reference}

@inproceedings{agarwal2024onpolicy,
  title = {On-Policy Distillation of Language Models: Learning from Self-Generated Mistakes},
  author = {Agarwal, Rishabh and Vieillard, Nino and Zhou, Yongchao and Stanczyk, Piotr and Ramos, Sabela and Geist, Matthieu and Bachem, Olivier},
  booktitle = {International Conference on Learning Representations},
  year = {2024}
}

@article{deepseek2025r1,
  title = {{DeepSeek-R1}: Incentivizing Reasoning Capability in {LLMs} via Reinforcement Learning},
  author = {{DeepSeek-AI}},
  journal = {Nature},
  volume = {645},
  pages = {633--638},
  year = {2025}
}

@misc{feng2026weaktostronggeneralizationdirectonpolicy,
  title = {Weak-to-Strong Generalization via Direct On-Policy Distillation},
  author = {Feng, Shiyuan and Gao, Huan-ang and Chi, Haohan and Wu, Hanlin and Zhang, Zhilong and Jiang, Zheng and He, Bingxiang and Ma, Wei-Ying and Zhang, Ya-Qin and Zhou, Hao},
  year = {2026},
  eprint = {2607.05394},
  archivePrefix = {arXiv},
  primaryClass = {cs.LG},
  url = {https://arxiv.org/abs/2607.05394}
}

@article{fu2023specializing,
  title = {Specializing Smaller Language Models towards Multi-Step Reasoning},
  author = {Fu, Yao and Peng, Hao and Ou, Litu and Sabharwal, Ashish and Khot, Tushar},
  journal = {arXiv preprint arXiv:2301.12726},
  year = {2023}
}

@inproceedings{gu2024minillm,
  title = {{MiniLLM}: Knowledge Distillation of Large Language Models},
  author = {Gu, Yuxian and Dong, Li and Wei, Furu and Huang, Minlie},
  booktitle = {International Conference on Learning Representations},
  year = {2024}
}

@misc{heo2026onpolicydeltadistillation,
  title = {On-Policy Delta Distillation},
  author = {Heo, Byeongho and Hwang, Jaehui and Yun, Sangdoo and Han, Dongyoon},
  year = {2026},
  eprint = {2607.15161},
  archivePrefix = {arXiv},
  primaryClass = {cs.LG},
  url = {https://arxiv.org/abs/2607.15161}
}

@misc{xu2026tip,
  title = {{TIP}: Token Importance in On-Policy Distillation},
  author = {Xu, Yuanda and Sang, Hejian and Zhou, Zhengze and He, Ran and Wang, Zhipeng and Geramifard, Alborz},
  year = {2026},
  eprint = {2604.14084},
  archivePrefix = {arXiv},
  primaryClass = {cs.LG},
  url = {https://arxiv.org/abs/2604.14084}
}

@misc{xu2026paced,
  title = {{PACED}: Distillation and On-Policy Self-Distillation at the Frontier of Student Competence},
  author = {Xu, Yuanda and Sang, Hejian and Zhou, Zhengze and He, Ran and Wang, Zhipeng},
  year = {2026},
  eprint = {2603.11178},
  archivePrefix = {arXiv},
  primaryClass = {cs.LG},
  url = {https://arxiv.org/abs/2603.11178}
}

@misc{wang2026tokenteachability,
  title = {Not All Disagreement Is Learnable: Token Teachability in On-Policy Distillation},
  author = {Wang, Yuanyi and Lu, Su and Gu, Yanggan and Wang, Pengkai and Yang, Yifan and Yan, Zhaoyi and Xie, Congkai and Wu, Jianmin and Yang, Hongxia},
  year = {2026},
  eprint = {2605.26844},
  archivePrefix = {arXiv},
  primaryClass = {cs.LG},
  url = {https://arxiv.org/abs/2605.26844}
}

@misc{shen2026inputspecificcredit,
  title = {From Generic Correlation to Input-Specific Credit in On-Policy Self-Distillation},
  author = {Shen, Guobin and Huang, Lei and Cheng, Xiang and Zhao, Chenxiao and Li, Jindong and Zhao, Dongcheng and Yu, Xing},
  year = {2026},
  eprint = {2605.11613},
  archivePrefix = {arXiv},
  primaryClass = {cs.LG},
  url = {https://arxiv.org/abs/2605.11613}
}

@misc{he2026selfdistillationzero,
  title = {Self-Distillation Zero: Self-Revision Turns Binary Rewards into Dense Supervision},
  author = {He, Yinghui and Kaur, Simran and Bhaskar, Adithya and Yang, Yongjin and Liu, Jiarui and Ri, Narutatsu and Fowl, Liam and Panigrahi, Abhishek and Chen, Danqi and Arora, Sanjeev},
  year = {2026},
  eprint = {2604.12002},
  archivePrefix = {arXiv},
  primaryClass = {cs.LG},
  url = {https://arxiv.org/abs/2604.12002}
}

@misc{li2026samplerouting,
  title = {Unifying Group-Relative and Self-Distillation Policy Optimization via Sample Routing},
  author = {Li, Gengsheng and Yang, Tianyu and Fang, Junfeng and Song, Mingyang and Zheng, Mao and Guo, Haiyun and Zhang, Dan and Wang, Jinqiao and Chua, Tat-Seng},
  year = {2026},
  eprint = {2604.02288},
  archivePrefix = {arXiv},
  primaryClass = {cs.LG},
  url = {https://arxiv.org/abs/2604.02288}
}

@misc{ye2025blackboxopd,
  title = {Black-Box On-Policy Distillation of Language Models},
  author = {Ye, Tianzhu and Dong, Li and Chi, Zewen and Wu, Xun and Huang, Shaohan and Wei, Furu},
  year = {2025},
  eprint = {2511.10643},
  archivePrefix = {arXiv},
  primaryClass = {cs.LG},
  url = {https://arxiv.org/abs/2511.10643}
}

@misc{sang2026crisp,
  title = {{CRISP}: Compressed Reasoning via Iterative Self-Policy Distillation},
  author = {Sang, Hejian and Xu, Yuanda and Zhou, Zhengze and He, Ran and Wang, Zhipeng and Sun, Jiachen},
  year = {2026},
  eprint = {2603.05433},
  archivePrefix = {arXiv},
  primaryClass = {cs.LG},
  url = {https://arxiv.org/abs/2603.05433}
}

@article{hinton2015distilling,
  title = {Distilling the Knowledge in a Neural Network},
  author = {Hinton, Geoffrey and Vinyals, Oriol and Dean, Jeff},
  journal = {arXiv preprint arXiv:1503.02531},
  year = {2015}
}

@inproceedings{hsieh2023distilling,
  title = {Distilling Step-by-Step! {O}utperforming Larger Language Models with Less Training Data and Smaller Model Sizes},
  author = {Hsieh, Cheng-Yu and Li, Chun-Liang and Yeh, Chih-Kuan and Nakhost, Hootan and Fujii, Yasuhisa and Ratner, Alexander and Krishna, Ranjay and Lee, Chen-Yu and Pfister, Tomas},
  booktitle = {Findings of the Association for Computational Linguistics: ACL 2023},
  year = {2023}
}

@article{li2022explanations,
  title = {Explanations from Large Language Models Make Small Reasoners Better},
  author = {Li, Shiyang and Chen, Jianshu and Shen, Yelong and Chen, Zhiyu and Zhang, Xinlu and Li, Zekun and Wang, Hong and Qian, Jing and Peng, Baolin and Mao, Yi and Chen, Wenhu and Xie, Xin},
  journal = {arXiv preprint arXiv:2210.06726},
  year = {2022}
}

@article{li2026rethinkopd,
  title = {Rethinking On-Policy Distillation of Large Language Models: Phenomenology, Mechanism, and Recipe},
  author = {Li, Yaxuan and Zuo, Yuxin and He, Bingxiang and Zhang, Jinqian and Xiao, Chaojun and Qian, Cheng and Yu, Tianyu and Gao, Huan-ang and Yang, Wenkai and Liu, Zhiyuan and Ding, Ning},
  journal = {arXiv preprint arXiv:2604.13016},
  year = {2026}
}

@inproceedings{liu2024ddk,
  title = {{DDK}: Distilling Domain Knowledge for Efficient Large Language Models},
  author = {Liu, Jiaheng and Zhang, Chenchen and Guo, Jinyang and Zhang, Yuanxing and Que, Haoran and Deng, Ken and Bai, Zhiqi and Liu, Jie and Zhang, Ge and Wang, Jiakai and Wu, Yanan and Liu, Congnan and Su, Wenbo and Wang, Jiamang and Qu, Lin and Zheng, Bo},
  booktitle = {Advances in Neural Information Processing Systems},
  year = {2024}
}

@article{fang2026rubricopd,
  title = {Rubric-based On-policy Distillation},
  author = {Fang, Junfeng and Hong, Zhepei and Zheng, Mao and Song, Mingyang and Li, Gengsheng and Jiang, Houcheng and Zhang, Dan and Guo, Haiyun and Wang, Xiang and Chua, Tat-Seng},
  journal = {arXiv preprint arXiv:2605.07396},
  year = {2026}
}

@article{hou2026uniopd,
  title = {{Uni-OPD}: Unifying On-Policy Distillation with a Dual-Perspective Recipe},
  author = {Hou, Wenjin and Peng, Shangpin and Wang, Weinong and Ruan, Zheng and Zhang, Yue and Zhou, Zhenglin and Gao, Mingqi and Chen, Yifei and Wang, Kaiqi and Yang, Hongming and Zhang, Chengquan and Tian, Zhuotao and Hu, Han and Yang, Yi and Wu, Fei and Fan, Hehe},
  journal = {arXiv preprint arXiv:2605.03677},
  year = {2026}
}

@article{song2026opdsurvey,
  title = {A Survey of On-Policy Distillation for Large Language Models},
  author = {Song, Mingyang and Zheng, Mao},
  journal = {arXiv preprint arXiv:2604.00626},
  year = {2026}
}

@article{singh2024beyond,
  title = {Beyond Human Data: Scaling Self-Training for Problem-Solving with Language Models},
  author = {Singh, Avi and Co-Reyes, John D. and Agarwal, Rishabh and Anand, Ankesh and Patil, Piyush and Garcia, Xavier and Liu, Peter J. and Harrison, J. and Lee, Jaehoon and Xu, Kelvin and Parisi, Aaron and Kumar, Abhishek and Alemi, Alex and Rizkowsky, Alex and Nova, Azade and Adlam, Ben and Bohnet, Bernd and Elsayed, Gamaleldin and Sedghi, Hanie and Mordatch, Igor and Simpson, Isabelle and G{"u}r, Izzeddin and Snoek, Jasper and Pennington, Jeffrey and Hron, Jiri and Kenealy, Kathleen and Swersky, Kevin and Mahajan, Kshiteej and Culp, Laura and Xiao, Lechao and Bileschi, Maxwell L. and Constant, Noah and Novak, Roman and Liu, Rosanne and Warkentin, Tris and Qian, Yundi and Bansal, Yamini and Dyer, Ethan and Neyshabur, Behnam and Sohl-Dickstein, Jascha and Fiedel, Noah},
  journal = {Transactions on Machine Learning Research},
  year = {2024}
}

@misc{xu2026grpoonpolicydistillationempirical,
  title = {Beyond {GRPO} and On-Policy Distillation: An Empirical Sparse-to-Dense Reward Principle for Language-Model Post-Training},
  author = {Xu, Yuanda and Sang, Hejian and Zhou, Zhengze and He, Ran and Wang, Zhipeng and Geramifard, Alborz},
  year = {2026},
  eprint = {2605.12483},
  archivePrefix = {arXiv},
  primaryClass = {cs.LG},
  url = {https://arxiv.org/abs/2605.12483}
}

@misc{yang2026learningteachergeneralizedonpolicy,
  title = {Learning beyond Teacher: Generalized On-Policy Distillation with Reward Extrapolation},
  author = {Yang, Wenkai and Liu, Weijie and Xie, Ruobing and Yang, Kai and Yang, Saiyong and Lin, Yankai},
  year = {2026},
  eprint = {2602.12125},
  archivePrefix = {arXiv},
  primaryClass = {cs.LG},
  url = {https://arxiv.org/abs/2602.12125}
}

@misc{ye2026opcd,
  title = {On-Policy Context Distillation for Language Models},
  author = {Ye, Tianzhu and Dong, Li and Wu, Xun and Huang, Shaohan and Wei, Furu},
  year = {2026},
  eprint = {2602.12275},
  archivePrefix = {arXiv},
  primaryClass = {cs.LG},
  url = {https://arxiv.org/abs/2602.12275}
}

@misc{ma2026mopdmultiteacheronpolicydistillation,
  title = {{MOPD}: Multi-Teacher On-Policy Distillation for Capability Integration in {LLM} Post-Training},
  author = {Ma, Wenhan and Wei, Jianyu and Zhao, Liang and Zhang, Hailin and Xiao, Bangjun and Li, Lei and Yang, Qibin and Gao, Bofei and Wang, Yudong and Li, Rang and Dong, Jinhao and Sui, Zhifang and Luo, Fuli},
  year = {2026},
  eprint = {2606.30406},
  archivePrefix = {arXiv},
  primaryClass = {cs.CL},
  url = {https://arxiv.org/abs/2606.30406}
}

@misc{chen2026counteractionawaremultiteacheronpolicydistillation,
  title = {Counteraction-Aware Multi-Teacher On-Policy Distillation for General Capability Recovery with Domain Preservation},
  author = {Chen, Tianlei and Ou, Jiao and Liu, Ziyuan and Tang, Ruiming and Liang, Jian and Li, Han},
  year = {2026},
  eprint = {2605.27115},
  archivePrefix = {arXiv},
  primaryClass = {cs.AI},
  url = {https://arxiv.org/abs/2605.27115}
}

@misc{li2026everycoin,
  title = {Every Coin Has Two Sides: On the Dual Nature of Generalization in On-Policy Distillation of Large Language Models},
  author = {Li, Zhaoyi and Kong, Deyang and Wei, Yuan and Yang, Evan and Shen, Ranran and Ihsani, Mahardika Krisna and Yang, Ming and Zhang, Wei and Hao, Chuan and Yang, Jian and Tao, Ran and Dai, Bryan and Zhang, Shikun and Ye, Wei and Wei, Ying and Lian, Defu},
  year = {2026},
  eprint = {2608.16647},
  archivePrefix = {arXiv},
  primaryClass = {cs.CL},
  url = {https://arxiv.org/abs/2608.16647}
}

@misc{shen2026diagnosingcalibratingtoolcallboundary,
  title = {Diagnosing and Calibrating Tool-Call Boundary Drift in Multi-Teacher On-Policy Distillation},
  author = {Shen, Jiabin and Chen, Guang and Mao, Chengjun},
  year = {2026},
  eprint = {2607.07050},
  archivePrefix = {arXiv},
  primaryClass = {cs.CL},
  url = {https://arxiv.org/abs/2607.07050}
}

@misc{lian2026uimopdmultiplatformonpolicydistillation,
  title = {{UI-MOPD}: Multi-Platform On-Policy Distillation for Continual {GUI} Agent Learning},
  author = {Lian, Niu and Chen, Alan and Yu, Zhehao and Duan, Chengzhen and Liu, Fazhan and Liu, Hui and Fu, Pei and Luan, Jian and Wang, Yaowei and Xia, Shu-Tao and Wang, Jinpeng},
  year = {2026},
  eprint = {2607.04425},
  archivePrefix = {arXiv},
  primaryClass = {cs.CL},
  url = {https://arxiv.org/abs/2607.04425}
}

@misc{chen2026regenreplayrecyclingexperttogeneralistdistillation,
  title = {{REGEN}: Replay-recycling for Expert-to-Generalist Distillation with Offline Reinforcement Learning},
  author = {Chen, Yunjie and Chen, Xiaoxin and Wang, Fang},
  year = {2026},
  eprint = {2607.19450},
  archivePrefix = {arXiv},
  primaryClass = {cs.LG},
  url = {https://arxiv.org/abs/2607.19450}
}

@inproceedings{ilharco2023task,
  title = {Editing Models with Task Arithmetic},
  author = {Ilharco, Gabriel and Ribeiro, Marco Tulio and Wortsman, Mitchell and
    Gururangan, Suchin and Schmidt, Ludwig and Hajishirzi, Hannaneh and
    Farhadi, Ali},
  booktitle = {International Conference on Learning Representations},
  year = {2023}
}

@inproceedings{liu2024proxy,
  title = {Tuning Language Models by Proxy},
  author = {Liu, Alisa and Han, Xiaochuang and Wang, Yizhong and Tsvetkov,
    Yulia and Choi, Yejin and Smith, Noah A.},
  booktitle = {International Conference on Learning Representations},
  year = {2024}
}

\appendix
\renewcommand{\dbltopfraction}{0.95}
\renewcommand{\topfraction}{0.95}
\renewcommand{\textfraction}{0.05}
\renewcommand{\dblfloatpagefraction}{0.8}
\setcounter{dbltopnumber}{6}
\setcounter{topnumber}{4}

\section{Protocols and Reproducibility}
\label{app:implementation-details}

\begin{table*}[t]
\centering
\small
\caption{Training protocols. Values are shared by both arms of a run
unless an arm-specific value is shown. Fields marked ``--'' were not
recorded in the run's frozen configuration summary.}
\label{tab:as-run-training}
\setlength{\tabcolsep}{2pt}
\begin{tabular}{@{}>{\raggedright\arraybackslash}p{0.13\textwidth}>{\raggedright\arraybackslash}p{0.15\textwidth}>{\raggedright\arraybackslash}p{0.15\textwidth}>{\raggedright\arraybackslash}p{0.13\textwidth}>{\raggedright\arraybackslash}p{0.16\textwidth}>{\raggedright\arraybackslash}p{0.2\textwidth}@{}}
\toprule
Setting & Acquisition ($M=1$) & Mechanism ($M=2$) & Scaling ($M=2,3$; proj.\ $M=4$)
& Phased routing & Interleaved routing\\
\midrule
Phase structure & single & single & single
& 2 phases, optimizer resumed, dataloader reset
& single, domains interleaved from step 1\\
On-policy prompts & BigMath, $25$K Math
& BigMath, $25$K Math & BigMath Math
& BigMath; TextbookReasoning $+$ Llama-Nemotron IF
& Math $+$ Science/IF mixture ($50/25/25$)\\
Temperature / top-$p$ & 1.0 / 1.0 & 1.0 / 1.0 & 1.0 / 1.0 & 1.0 / 1.0
& 1.0 / 1.0, top-$k$ off\\
Maximum response & 10{,}000 & 10{,}000 & 10{,}000 & 10{,}000 & 10{,}000\\
Learning rate & $10^{-6}$ & $10^{-6}$ & $10^{-6}$ &$10^{-6}$ 
& $10^{-6}$\\
Hardware & 4 $\times$ H100 80GB & 4 $\times$ H100 80GB & 8 $\times$ H200
& 4 $\times$ H100 80GB per arm
& 8 $\times$ H200 (Endpoint) / 8 $\times$ H100 (\method)\\
Evaluation & Avg@$K$ (Table~\ref{tab:as-run-evaluation})
& Avg@$K$ and greedy & greedy pass@$1$ & greedy pass@$1$
& greedy pass@$1$\\
\bottomrule
\end{tabular}
\end{table*}

\begin{table*}[t]
\centering
\small
\caption{Held-out evaluation suite. Avg@$K$ applies to the acquisition
and mechanism runs; other runs decode every benchmark greedily at
$K=1$.}
\label{tab:as-run-evaluation}
\resizebox{\textwidth}{!}{%
\begin{tabular}{llrrl}
\toprule
Group & Benchmark & Test items & Avg@$K$ & Generation / scoring\\
\midrule
English Math & AMC 2023 & 40 & 16 & sampled, official answer checker\\
English Math & MATH-500 & 500 & 1 & greedy, official answer checker\\
English Math & AIME 2025 & 30 & 16 & sampled, official answer checker\\
Science & GPQA-Diamond & 198 & 4 & sampled, final option-letter scorer\\
Instruction following & IF-Eval & 541 & 4 & sampled, vendored rule
checker\\
\bottomrule
\end{tabular}
}
\end{table*}

\paragraph{Evaluation.}
Evaluation uses SGLang with a $16{,}384$-token maximum response length,
except that IF-Eval is capped at $2{,}048$ tokens by the evaluation
driver; the scaling run and teacher references use a $10{,}000$-token
greedy protocol. Greedy decoding uses temperature $0$ and top-$p=1$;
Avg@$K$ sampling uses temperature $1$ and top-$p=0.95$. MATH-500 reuses
its greedy item outcomes for $K=1$. All runs use the same $1{,}309$
frozen held-out items. Five-suite macros are unweighted averages of the
five benchmark scores, not of the two group macros. Entries with error
bars are mean $\pm$ sample standard deviation over five independently
trained seeds with matched seed IDs and prompt orders; macros are
computed from the five-seed benchmark means.

\paragraph{Reproducibility.}
Each teacher is paired with its exact precursor and checked for
tokenizer compatibility; the Qwen3 extension records the token-string
overlap map of Eq.~\ref{eq:qwen3-projected-shift}. Paired arms share
prompts, update counts, and evaluation settings, and all frozen models
score identical student prefixes within an arm before target
construction. Dataset revisions, sampled IDs, chat templates, model
revisions, decoding settings, and verifier versions are fixed across
comparisons.

\section{Estimator and Diagnostic Definitions}
\label{app:diagnostic-definitions}

\paragraph{Estimator and normalization.}
For valid sampled-token positions $\mathcal I$ in an optimization batch,
training subtracts the empirical mean reward
$\bar R^C=|\mathcal I|^{-1}\sum_{(b,t)\in\mathcal I}R^C_{b,t}(y_{b,t})$
and optimizes the detached centered reward times the student
log-probability. This is sampled-token centering rather than an explicit
per-state full-vocabulary expectation. When a restricted support $S_t$ is
used, the sampled-token numerator is exact and only the composite
partition is approximated:
\begin{equation}
\begin{aligned}
\widehat{\log\pi_C}(y_t\mid h_t)
={}&z_C(y_t\mid h_t)\\
&-\log\sum_{u\in S_t}\exp z_C(u\mid h_t).
\end{aligned}
\label{eq:support-normalizer}
\end{equation}
If $m_{S_t}=\sum_{v\in S_t}\pi_C(v\mid h_t)$ is the composite mass
captured by the support, then
$\widehat{\log\pi_C}(y_t\mid h_t)=\log\pi_C(y_t\mid h_t)-\log m_{S_t}$,
so the restricted normalizer raises the sampled-token log probability by
$-\log m_{S_t}$. This is a normalizer approximation, not the top-$k$ OPD
objective. All logits are restricted to the $151{,}665$-token effective
vocabulary before composition.

\paragraph{Gradient decomposition.}
For sampled positions $\{(h_t,y_t)\}_{t=1}^{N}$, define
$R^\Delta_t=\log\pi_T(y_t\mid h_t)-\log\pi_B(y_t\mid h_t)$ and
$R^B_t=\log\pi_B(y_t\mid h_t)-\log\pi_\theta(y_t\mid h_t)$. Their
gradient estimators and relative magnitude are
\begin{equation}
\begin{aligned}
g_\bullet
&=\nabla_\theta\!\left[
-\frac{1}{N}\sum_{t=1}^{N}
\sg\!\big(R^\bullet_t\big)\,\log\pi_\theta(y_t\mid h_t)
\right],\\
&\qquad\bullet\in\{\Delta,B\},\\
\Gamma_{\mathrm{base}}
&=\frac{\|g_B\|_2}{\|g_\Delta\|_2+10^{-8}}.
\end{aligned}
\label{eq:gradient-components}
\end{equation}
At each mechanism checkpoint, Eq.~\ref{eq:gradient-components} uses the
cross-origin teacher--base pair, the first micro-batch, and its first
$N=\min(128,|\mathcal I|)$ valid response positions, backpropagating
through all trainable parameters. Measurements are recorded every $20$
steps from step $20$ through $380$.

\paragraph{Composition geometry.}
Center each shift by its vocabulary mean,
\begin{equation}
\begin{aligned}
\Delta_i(t,v)
={}&z_{T_i}(v\mid h_t)-z_{B_i}(v\mid h_t)\\
&-\frac{1}{V}\sum_{u=1}^{V}
\left[z_{T_i}(u\mid h_t)-z_{B_i}(u\mid h_t)\right],
\end{aligned}
\label{eq:centered-shift}
\end{equation}
concatenate across sampled positions, and compute pairwise cosine,
aggregate norm, and
\begin{equation}
\mathrm{Cancel}(\mathcal T)
=1-
\frac{\left\|\sum_{i\in\mathcal T}\Delta_i\right\|_2}
{\sum_{i\in\mathcal T}\|\Delta_i\|_2+10^{-8}}.
\label{eq:cancellation}
\end{equation}
Sign conflict uses the top-$16$ absolute-shift coordinates selected by
each teacher, retaining overlaps and aggregating across positions. The
endpoint control replaces $z_{T_i}-z_{B_i}$ by $z_{T_i}-z_A$. Because
$\Delta_i$ is evaluated at student-visited prefixes, these statistics
depend on the rollout protocol and pooling rule. The mechanism run
concatenates the first $128$ valid positions of the first micro-batch on
each data-parallel rank; the scaling run computes the same quantities per
complete response and averages over micro-batches. We read them within a
run.

Two orthogonal equal-norm vectors have cosine $0$ and
$\mathrm{Cancel}=1-\sqrt2/2=0.2929$; independent sign-symmetric
coordinates have conflict rate $0.5$. For
$a=\Delta_{\mathrm{same}}$, $b=\Delta_{\mathrm{cross}}$,
$c=\cos(a,b)$, and $r=\|b\|/\|a\|$,
\begin{equation}
1-\mathrm{Cancel}
=\frac{\sqrt{r^2+1+2rc}}{r+1}.
\label{eq:norm-ratio}
\end{equation}
Since this expression is invariant under $r\mapsto1/r$, cosine and
cancellation identify the norm ratio $\max(r,1/r)$.

\paragraph{Target distance.}
The target-to-student forward KL is
\begin{equation}
D_{\mathrm{ST}}
=\mathbb E_{h_t}\sum_{v=1}^{V}
\pi_C(v\mid h_t)
\log\frac{\pi_C(v\mid h_t)}{\pi_\theta(v\mid h_t)},
\label{eq:student-target-kl}
\end{equation}
computed over the full $V=151{,}665$ vocabulary on the first micro-batch
of each mechanism checkpoint, using up to $128$ valid response positions.
KL to the anchor replaces $\pi_\theta$ by $\pi_A$.

\section{Common-Domain Composition Details}
\label{app:composition-details}

\subsection{Mechanism run: capability}
\label{app:k2-greedy}

\begin{table*}[t]
\centering
\small
\caption{Mechanism run ($M=2$) student evaluation. AMC 2023 and
AIME 2025 use Avg@$16$, GPQA-Diamond and IF-Eval use Avg@$4$, and
MATH-500 uses greedy decoding. Entries with error bars are mean $\pm$
sample standard deviation over training seeds, in percent.}
\label{tab:k2-capability}
\resizebox{\textwidth}{!}{%
\begin{tabular}{lcccc}
\toprule
Benchmark & Initial student & \method & Endpoint composite
& \method\ $-$ Endpoint (pp) \\
\midrule
AMC 2023 & 63.12 & $79.84 \pm 1.99$
& $79.53 \pm 2.79$ & $+0.31$\\
MATH-500 & 36.00 & $59.40 \pm 1.29$
& $62.80 \pm 1.34$ & $-3.40$\\
AIME 2025 & 23.33 & $32.92 \pm 3.04$
& $32.92 \pm 3.24$ & $0.00$\\
\textbf{English Math macro} & \textbf{40.82} & \textbf{57.39}
& \textbf{58.42} & $\mathbf{-1.03}$\\
GPQA-Diamond & 34.72 & $40.28 \pm 2.11$
& $39.02 \pm 2.22$ & $+1.26$\\
IF-Eval & 23.43 & $24.58 \pm 1.31$
& $24.68 \pm 1.07$ & $-0.10$\\
\textbf{Science/IF macro} & \textbf{29.08} & \textbf{32.43}
& \textbf{31.85} & $\mathbf{+0.58}$\\
\bottomrule
\end{tabular}
}
\end{table*}

Under greedy decoding, the group-balanced totals are $24.10\%$ for the
initial student, $37.73\%$ for \method, and $36.25\%$ for the endpoint
composite. The per-group greedy macros are $31.44\%$, $50.91\%$, and
$48.16\%$ for English-Math, and $16.76\%$, $24.54\%$, and $24.34\%$ for
Science/IF, so greedy decoding favors \method\ on both groups while the
Avg@$K$ ordering is effectively tied.

\subsection{Mechanism run: diagnostics}
\label{app:k2-early-mechanism}

\begin{table*}[t]
\centering
\small
\caption{Late-training diagnostics for the mechanism run, averaged over
steps $300,320,340,360,380$.}
\label{tab:k2-composition-diagnostics}
\resizebox{\textwidth}{!}{%
\begin{tabular}{lccccccc}
\toprule
Target & Cancellation & Aggregate norm & Conflict rate & Signal cosine
& Entropy & KL to anchor & Target--student KL\\
\midrule
\method
& 0.2871 & 3247.5 & 0.5540 & $-0.0163$
& 0.2458 & 0.1243 & 0.0307\\
Endpoint composite
& 0.1363 & 10231.5 & 0.4386 & 0.0592
& 0.2643 & 0.2567 & 0.1522\\
\midrule
\emph{Equal-norm reference}
& 0.2929 & --- & 0.5000 & 0.0000 & --- & --- & ---\\
\bottomrule
\end{tabular}
}
\end{table*}

\begin{table}[t]
\centering
\small
\caption{Early mechanism-run diagnostics at steps $20, 60, 100$.}
\label{tab:k2-early-mechanism}
\resizebox{\columnwidth}{!}{%
\begin{tabular}{lrrr}
\toprule
Target & Step & Cancellation & Target--student KL\\
\midrule
\method & 20  & 0.2728 & 0.1070\\
\method & 60  & 0.2914 & 0.0750\\
\method & 100 & 0.2832 & 0.0462\\
Endpoint composite & 20  & 0.1265 & 0.2472\\
Endpoint composite & 60  & 0.1378 & 0.1222\\
Endpoint composite & 100 & 0.1381 & 0.1632\\
\midrule
\emph{Reference} & --- & 0.2929 & ---\\
\bottomrule
\end{tabular}
}
\end{table}

Table~\ref{tab:k2-composition-diagnostics} reports the late-training
window and Table~\ref{tab:k2-early-mechanism} early checkpoints; the two
arms separate by step $20$ and the ordering is stable.

\subsection{Scaling run}
\label{app:m3-scaling}

\begin{table*}[t]
\centering
\small
\caption{Scaling run at step $100$. Greedy pass@$1$; entries with error
bars are mean $\pm$ sample standard deviation over training seeds, in
percent.}
\label{tab:m3-full-capability}
\setlength{\tabcolsep}{4pt}
\resizebox{\textwidth}{!}{%
\begin{tabular}{@{}llrrrrrrrr@{}}
\toprule
Teachers & Target & AMC 2023 & MATH-500 & AIME 2025 & Math macro
& GPQA-D & IF-Eval & Sci/IF macro & Five-suite macro\\
\midrule
$M=2$ & Endpoint composite & $35.50 \pm 2.56$
& $34.04 \pm 1.37$ & $17.33 \pm 1.93$ & 28.96
& $16.77 \pm 2.27$ & $14.64 \pm 1.66$ & 15.70 & 23.66\\
$M=2$ & \method & $38.00 \pm 1.68$
& $28.64 \pm 1.42$ & $20.67 \pm 3.11$ & 29.10
& $20.30 \pm 1.90$ & $14.45 \pm 1.33$ & 17.38 & 24.41\\
$M=3$ & Endpoint composite & $38.00 \pm 1.80$
& $32.84 \pm 0.99$ & $17.33 \pm 3.08$ & 29.39
& $17.27 \pm 1.53$ & $15.75 \pm 1.14$ & 16.51 & 24.24\\
$M=3$ & \method & $38.00 \pm 1.61$
& $31.84 \pm 0.99$ & $30.67 \pm 2.50$ & \textbf{33.50}
& $17.27 \pm 1.48$ & $13.16 \pm 1.50$ & 15.22
& \textbf{26.19}\\
\bottomrule
\end{tabular}
}
\end{table*}

Under the same greedy, $10{,}000$-token protocol, the individual teachers
score (Math\,/\,Sci-IF\,/\,five-suite macro, \%) $46.62/28.08/39.21$ for
Nemotron-1.5B, $49.02/27.50/40.41$ for JustRL-1.5B, and
$59.29/36.87/50.32$ for Polaris-7B.

At $M=2$ the two shifts have norms $13{,}674$ and $6{,}073$ with
pairwise cosine $0.084$. Adding JustRL gives norms $12{,}891$,
$11{,}114$, and $6{,}490$, so the third shift enters at a magnitude
comparable to the other two. Two of its pairwise cosines remain near zero
($0.072$ and $0.062$) while the third rises to $0.315$. Across the
recorded $M=2$ and $M=3$ training-signal diagnostics, shift composition
lowers target displacement, reward variance, and gradient magnitude; at
$M=3$, its final target-to-anchor KL is $0.483$ rather than $0.695$.
Its per-update time is $11.0\%$ higher ($167.1$ versus $150.6$ seconds)
because the Polaris precursor must also be scored.

\section{Routed-Domain Distillation Details}
\label{app:routed}

\paragraph{Interleaved routing.}
The router is the prompt's domain: Math prompts are supervised by
Polaris and Science and instruction-following prompts by Nemotron, with
exactly one teacher per prompt. The training mixture is $50\%$ Math,
$25\%$ Science, and $25\%$ instruction-following. Domains are
interleaved from the first update
with a deterministic within-batch shuffle. Endpoint matches
$z_{T_{\rho(h)}}$ and \method\ matches $z_A+\delta_{\rho(h)}$. Targets are
normalized exactly over the full effective vocabulary. The Endpoint arm
was trained on $8\times$H200 and the \method\ arm on
$8\times$H100; the comparison is at matched update count, so hardware
affects throughput but not the reported accuracies. All checkpoints were
evaluated on H100 with greedy decoding.

\subsection{Phased routing}
\label{app:seq-tests}

\begin{table*}[t]
\centering
\small
\caption{Phased routing at the step-$300$ checkpoint. Greedy pass@$1$;
group macros are unweighted averages within each group. Entries with
error bars are mean $\pm$ sample standard deviation over training seeds,
in percent.}
\label{tab:seq-capability}
\resizebox{\textwidth}{!}{%
\begin{tabular}{llccccccc}
\toprule
Order & Method & AMC 2023 & MATH-500 & AIME 2025 & GPQA-Diamond & IF-Eval
& Math macro & Sci/IF macro \\
\midrule
\multirow{2}{*}{Science/IF $\rightarrow$ Math}
& Endpoint & $25.50 \pm 2.43$ & $41.24 \pm 0.84$
& $17.33 \pm 2.19$ & $15.76 \pm 1.93$
& $20.74 \pm 0.81$ & 28.02 & 18.25\\
& \method  & $33.00 \pm 2.06$ & $42.84 \pm 1.31$
& $27.33 \pm 2.69$ & $24.85 \pm 2.28$
& $20.92 \pm 1.92$ & 34.39 & 22.89\\
\midrule
\multirow{2}{*}{Math $\rightarrow$ Science/IF}
& Endpoint & $40.50 \pm 1.79$ & $51.44 \pm 0.94$
& $30.67 \pm 2.21$ & $29.90 \pm 1.81$
& $20.55 \pm 1.35$ & 40.87 & 25.23\\
& \method  & $40.50 \pm 2.48$ & $55.44 \pm 1.12$
& $30.67 \pm 2.52$ & $32.42 \pm 2.43$
& $22.03 \pm 1.47$ & 42.20 & 27.23\\
\bottomrule
\end{tabular}
}
\end{table*}

Each arm trains two phases with actor and optimizer state carried across
the switch. The Science/IF phase is identical across methods, so the
contrast lies in the Polaris Math phase. Per-item binary outcomes are
retained, so same-order \method-versus-Endpoint comparisons are exact
paired McNemar tests on identical items. The only comparison below
$p=0.05$ is MATH-500 in Math $\to$ Science/IF ($56$ versus $36$
discordant items, $p=0.047$), which does not survive correction across
the ten benchmark-by-order comparisons.

\section{Cross-Origin Acquisition and Compute}
\label{app:k1-later}

\begin{table}[t]
\centering
\small
\caption{Acquisition run at step $100$. AMC 2023 and AIME 2025 use
Avg@$16$; MATH-500 uses greedy decoding. The macro is the unweighted
average of these three benchmarks. Entries with error bars are mean
$\pm$ sample standard deviation over training seeds, in percent.}
\label{tab:k1-cross-origin}
\resizebox{\columnwidth}{!}{%
\begin{tabular}{lcccc}
\toprule
Target & AMC 2023 & MATH-500 & AIME 2025 & English Math macro\\
\midrule
\method & $67.81 \pm 1.53$ & $48.80 \pm 1.40$
& $27.08 \pm 1.82$ & \textbf{47.90}\\
Endpoint & $64.84 \pm 1.71$ & $44.20 \pm 0.93$
& $26.04 \pm 1.73$ & 45.03\\
\bottomrule
\end{tabular}
}
\end{table}

Both arms train only on BigMath prompts. The English-Math macro across
the evaluated checkpoints is $47.90/45.65/47.68$ for \method\ and
$45.03/46.18/46.54$ for Endpoint at steps $100/150/195$. On GPQA-Diamond
and IF-Eval, the ordering flips with decoding mode: at step $100$,
\method\ leads by $0.91$ pp on the Avg@$4$ macro while greedy decoding
favors Endpoint by $1.07$ pp. The endpoint target remains near $5\times$
farther from the student in target--student KL across checkpoints.

\subsection{Compute accounting}
\label{app:compute-convergence}

For an arm with $S$ updates, measured end-to-end step times $t_s$, and
$N_{\mathrm{GPU}}$ H100 GPUs, allocated compute is
\begin{equation}
C_{\mathrm{alloc}}
=\frac{N_{\mathrm{GPU}}}{3600}\sum_{s=1}^{S}t_s
\quad\text{H100-hours}.
\label{eq:allocated-compute}
\end{equation}
Thus, one wall-clock hour of training on four allocated H100s counts as
four H100-hours. The timed region includes rollout generation, student
optimization, batching, and every teacher, precursor, and anchor forward
pass, but excludes held-out evaluation. This measures allocated device
time rather than FLOPs and does not correct for utilization. The $M=1$
run uses each arm's mean step time and four GPUs; the $M=2$ run sums its
recorded step times before multiplying by four GPUs.

For the acquisition comparison, \method{} first exceeds Endpoint's best
evaluated English-Math macro at step $100$, whereas Endpoint attains that
best score at step $195$. With mean step times of approximately $252$ and
$198$ seconds, respectively, their allocated costs are

\begin{equation}
\begin{aligned}
C_{\Delta\text{-MOPD}}&=4\times100\times252/3600=28.0,\\
C_{\mathrm{Endpoint}}&=4\times195\times198/3600=42.9
\end{aligned}
\quad\text{H100-hours}.
\label{eq:acquisition-compute-example}
\end{equation}
Therefore the matched-capability reduction is
$(42.9-28.0)/42.9=34.7\%$, reported as $35\%$. The reduction comes from
requiring fewer updates, not from cheaper updates: \method{} is $27.3\%$
slower per update because it also scores the precursor.

\begin{table}[t]
\centering
\small
\caption{Earliest evaluated checkpoint meeting each held-out English-Math
threshold in the acquisition run, in allocated H100-hours. Checkpoints
were evaluated every $50$ steps, so every entry is an upper bound on
first-passage compute.}
\label{tab:k1-isocapability}
\setlength{\tabcolsep}{4pt}
\begin{tabular}{@{}lcc@{}}
\toprule
English-Math macro & \method & Endpoint\\
\midrule
$46.18\%$ & $\le 28.0$ h & $\le 33.0$ h\\
$46.54\%$ & $\le 28.0$ h & $\le 42.9$ h\\
$47.90\%$ & $\le 28.0$ h & not observed by $42.9$ h\\
\bottomrule
\end{tabular}
\end{table}

At equal update count, \method\ costs $27.3\%$ more in $M=1$ and
$12.0\%$ more in $M=2$, below a naive factor of two because rollout
generation, optimization, batching, and parallel scoring are shared.
Every H100-hour figure, including Table~\ref{tab:k1-isocapability},
already charges \method\ for this overhead. In the separate scaling run,
the $M=3$ overhead is $11.0\%$ on the same eight-H200 configuration in
both arms; this is a within-run wall-clock ratio, not an H200-to-H100
conversion.

\paragraph{Convergence criterion.}
We define the plateau as the first $25$-step block after which every
later block-mean rollout accuracy varies by at most $2.5$ pp. \method\
stabilizes near step $75$ and Endpoint near step $125$, corresponding to
$21.0$ versus $27.5$ H100-hours ($1.31\times$). At $21.0$ hours \method\
has $53.38\%$ block-mean rollout accuracy, against $51.12\%$ for Endpoint
at a comparable $22.0$ hours. Best-macro compute, updates to plateau, and
accuracy at matched hours therefore agree.

\section{Cross-Tokenizer Extension}
\label{app:qwen3}

This extension adds Qwen3-DAPO-449, our public Qwen3-8B-DAPO iter-449
checkpoint, as a second cross-origin teacher in the scaling run ($M=4$).
It is tokenizer-identical to its Qwen3-8B precursor but not fully
special-token compatible with the DeepSeek/Qwen2 anchor.

\paragraph{Overlap projection.}
We compute the Qwen3 shift inside the Qwen3 family and transfer only
overlapping action coordinates to the anchor vocabulary. Let $V_A$ be the
anchor vocabulary, $V_Q$ the Qwen3 vocabulary, and $\tau_A,\tau_Q$ their
id-to-token-string maps. With
\begin{equation}
\begin{aligned}
\delta_Q^Q(v_Q\mid h_t)
={}&z_{\mathrm{DAPO}}^Q(v_Q\mid h_t)\\
&-z_{\mathrm{Qwen3\mbox{-}8B}}^Q(v_Q\mid h_t),
\end{aligned}
\end{equation}
define $m(v_A)=v_Q$ whenever $\tau_A(v_A)=\tau_Q(v_Q)$. The projected
shift is
\begin{equation}
\widetilde{\delta}_Q(v_A\mid h_t)=
\begin{cases}
\delta_Q^Q(m(v_A)\mid h_t), & m(v_A)\in V_Q,\\
0, & \text{otherwise},
\end{cases}
\label{eq:qwen3-projected-shift}
\end{equation}
and the $M=4$ target is
$z_A+\sum_{i=1}^{3}\delta_i+\widetilde{\delta}_Q$. We verified
$151{,}660$ token-string overlaps, no duplicate token strings, and
$151{,}658$ shared tokens with identical ids. Only \texttt{<think>} and
\texttt{</think>} require remapping; five anchor-only special tokens
receive zero Qwen3 shift, and Qwen3-only special tokens are dropped. This
full-vocabulary overlap projection is narrower than text-span likelihood
alignment for general cross-tokenizer OPD. Because cross-tokenizer
endpoint logits are not placed on the anchor vocabulary, no paired
endpoint control is defined.

\begin{table*}[t]
\centering
\small
\caption{Projected four-teacher extension at step $100$, with the
$M=3$ shift composite for reference. Greedy pass@$1$, in percent.}
\label{tab:qwen3-extension}
\setlength{\tabcolsep}{4pt}
\resizebox{\textwidth}{!}{%
\begin{tabular}{@{}llrrrrrrrr@{}}
\toprule
Teachers & Target & AMC 2023 & MATH-500 & AIME 2025 & Math macro
& GPQA-D & IF-Eval & Sci/IF macro & Five-suite macro\\
\midrule
$M=3$ & \method & $38.00 \pm 1.61$
& $31.84 \pm 0.99$ & $30.67 \pm 2.50$ & 33.50
& $17.27 \pm 1.48$ & $13.16 \pm 1.50$ & 15.22 & 26.19\\
$M=4$ & Projected \method & $40.50 \pm 1.92$
& $32.64 \pm 1.19$ & $37.33 \pm 2.63$ & 36.82
& $16.77 \pm 1.76$ & $12.94 \pm 1.42$ & 14.85 & 28.04\\
\bottomrule
\end{tabular}
}
\end{table*}

Adding the projected shift raises the Math macro from $33.50$ to $36.82$
and the five-suite macro from $26.19$ to $28.04$. All three Math
benchmarks improve, with the largest gain on AIME 2025, while Science/IF
does not improve. For reference, the
Qwen3-DAPO-449 model card
(\url{https://huggingface.co/pb09204048/Qwen3-8B-DAPO-iter449-disable-thinking})
reports AIME-24/AIME-25 pass@$1$ of $73.75\%/67.08\%$ under its
non-thinking sampling protocol ($n=8$, maximum response length
$30{,}000$), compared with $25.00\%/17.90\%$ for Qwen3-8B under the same
protocol.

\end{document}